\documentclass[10pt,twocolumn,letterpaper]{article}
\usepackage{cvpr}
\usepackage{times}
\usepackage{epsfig}
\usepackage{graphicx}
\usepackage{amsmath}
\usepackage{amssymb}

\usepackage{url}
\usepackage{bm}

\usepackage{enumitem}
\usepackage{comment}
\usepackage{amsmath}
\DeclareMathOperator*{\argmax}{arg\,max}

\usepackage{caption}
\usepackage{subcaption}
\usepackage{authblk}
\usepackage{lipsum}
\usepackage{algorithm}
\usepackage{algpseudocode}
\usepackage{lipsum}

\newcommand*{\affaddr}[1]{#1} 
\newcommand*{\affmark}[1][*]{\textsuperscript{#1}}

\newcommand\blfootnote[1]{%
  \begingroup
  \renewcommand\thefootnote{}\footnote{#1}%
  \addtocounter{footnote}{-1}%
  \endgroup
}

\usepackage[pagebackref=true,breaklinks=true,letterpaper=true,colorlinks,bookmarks=false]{hyperref}

\cvprfinalcopy 

\def\cvprPaperID{4603} 

\ifcvprfinal\fi
\begin{document}

\title{Multimodal Explanations by Predicting Counterfactuality in Videos}

\author{%
 Atsushi Kanehira\affmark[1], Kentaro Takemoto\affmark[2], Sho Inayoshi\affmark[2], and Tatsuya Harada\affmark[2,3]\\
\affaddr{\affmark[1]Preferred Networks, \affmark[2]The University of Tokyo}, \affaddr{\affmark[3]RIKEN}}

\maketitle
\thispagestyle{empty}

\begin{abstract}
This study addresses generating counterfactual explanations with multimodal information. Our goal is not only to classify a video into a specific category, but also to provide explanations on why it is not categorized to a specific class with combinations of visual-linguistic information.
Requirements that the expected output should satisfy are referred to as counterfactuality in this paper: (1) Compatibility of visual-linguistic explanations, and (2) Positiveness/negativeness for the specific positive/negative class. 
Exploiting a spatio-temporal region (tube) and an attribute as visual and linguistic explanations respectively, the explanation model is trained to predict the counterfactuality for possible combinations of multimodal information in a post-hoc manner. 
The optimization problem, which appears during training/inference, can be efficiently solved by inserting a novel neural network layer, namely the maximum subpath layer. We demonstrated the effectiveness of this method by comparison with a baseline of the action recognition datasets extended for this task. Moreover, we provide information-theoretical insight into the proposed method.
\end{abstract}

\blfootnote{This work is done at the University of Tokyo.}

\section{Introduction}
The visual cognitive ability of machines has significantly improved mostly due to the recent development of deep learning techniques. Owing to its high complexity, the decision process is inherently a black-box, and therefore, much research has focused on making the machine explain the reason behind its decision to verify its trustability.

The present study particularly focuses on building a system that not only classifies a video, but also explains why a given sample is {\bf not} predicted to one class but another, in spite of almost all existing research pursuing the reason for the positive class. Concretely, we intend to generate the explanation in the form ``X is classified to A not B because C and D exists in X.''\footnote{rephrased by ``X would be  classified as B not A if C and D not in X.''}
This type of explanation is referred to as {\it counterfactual explanation}~\cite{wachter2017counterfactual} in this paper. It may aid us in understanding why the model prediction is different from what we think, or when discrimination is difficult between two specific classes. 
The explanation is valuable especially for long videos because it provides information efficiently on the content of what humans cannot see immediately.

\begin{figure}[t!]
\begin{center}
\begin{tabular}{c}
\includegraphics[clip, width=0.97\linewidth, height=3.9cm]{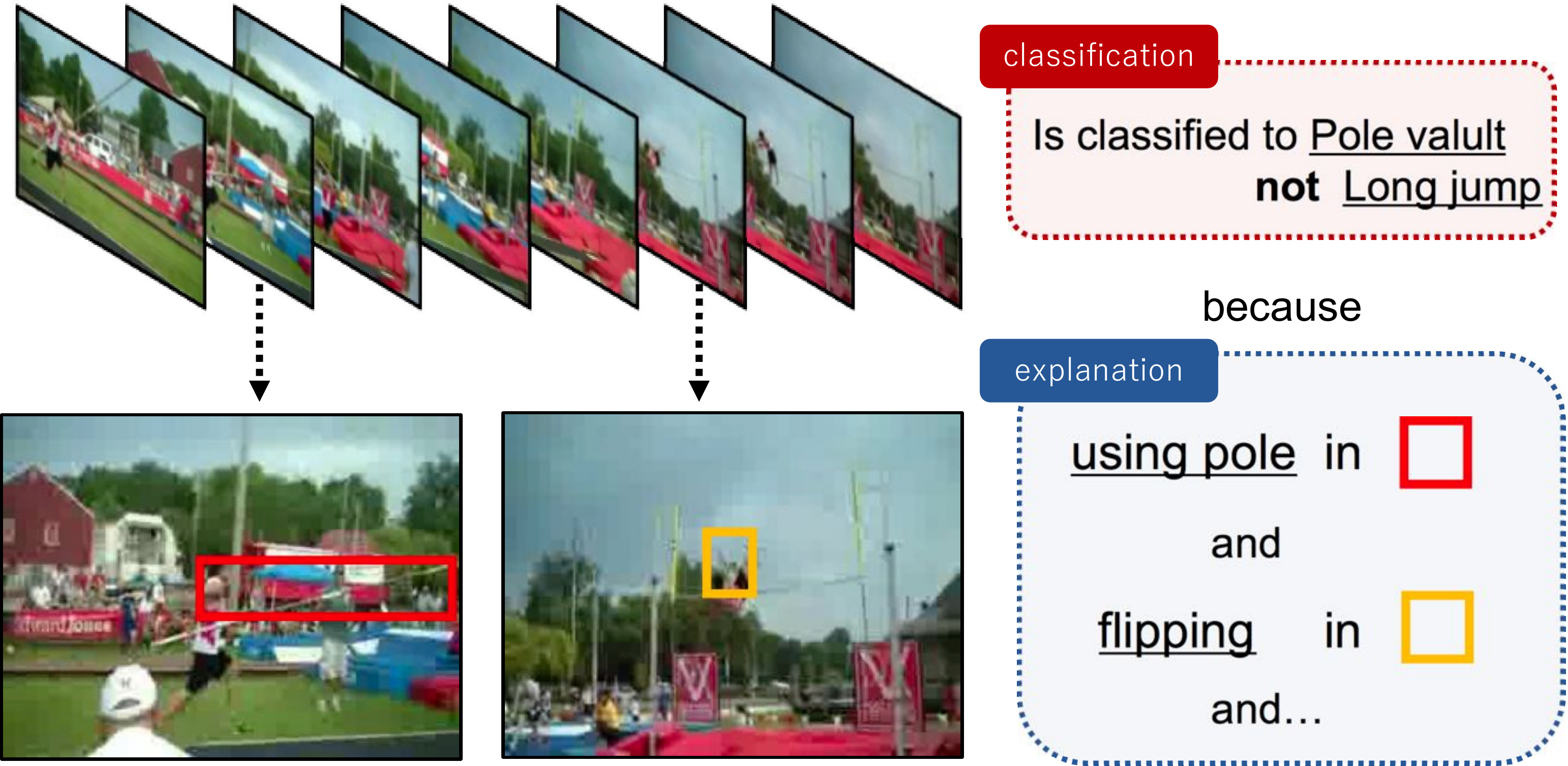}
\end{tabular}
\end{center}
\vspace{-0.35cm}
\caption{Our model not only classifies a video to a category (Pole vault), but also generates explanations why the video is not classified to another class (Long jump). It outputs several pairs of attribute (e.g., using pole) and spatio-temporal region (e.g., red box) as an explanation.}
\vspace{-0.45cm}
\label{fig:example}
\end{figure}

We first need to discuss the desired output. This work treats an explanation as the one satisfying two conditions as follows:
\begin{enumerate}[label=(\Alph*), noitemsep,topsep=0pt]
 \setlength{\itemsep}{0pt}
  \setlength{\parskip}{0pt}
\item The output should be interpretable for humans,
\item The output should have fidelity to the explained target.
\end{enumerate}

Related to (A), one natural way of obtaining interpretable output would be to assign an importance to each element in the input space and visualize it. Although this can help us to perceive the important region for the model prediction, the interpretation leading to it cannot be uniquely determined. We enhance the interpretability of the explanation by leveraging not only parts of visual input but also linguistics, which is compatible with the visual information similar to the previous work~\cite{Hendricks_2018_ECCV}.
More specifically, dealing with a spatio-temporal region of the target video and (the existence of) an attribute as elements, we concatenate them to generate explanations.
An example is shown in the Fig.~\ref{fig:example}.

To realize (B) while satisfying (A), the expected output of the visual-linguistic explanation should have the following two properties: 
\begin{enumerate}[noitemsep,topsep=0pt]
\setlength{\itemsep}{0pt}
  \setlength{\parskip}{0pt}
\item[(1)] Visual explanation is the region which retains high positiveness/negativeness on model prediction for specific positive/negative classes,
\item[(2)] Linguistic explanation is compatible with the visual counterpart. 
\end{enumerate}
The score to measure how the requirements above are fulfilled is hereafter referred to as the {\it counterfactuality score}, or simply {\it counterfactuality}.

The above-listed requirements cannot be achieved by naively exploiting output of the existing method, which considers only positiveness for explanation, such as in ~\cite{Hendricks_2018_ECCV}, where the authors attempt to generate visual-linguistic explanations of positive class.
This is mainly because positiveness/negativeness need to be considered simultaneously for specific positive/negative classes in the same region.

To build a system that generates explanations satisfying both (1) and (2), we propose a novel framework for generating counterfactual explanations based on predicting counterfactuality. The outline of the framework is depicted in two steps:
(a) Train a classification model that is the target of the explanation,
(b) Train an auxiliary explanation model in a post-hoc manner by utilizing output and mid-level features of the target classifier after freezing its weights to prevent output change.
An explanation model predicts counterfactuality scores for all negative classes.
It is trained by exploiting the fact that supervised information ``X is classified to category A.'' can be translated into ``X is not classified to any category B except A.'' 

The proposed explanation model holds a trainable classifier that predicts simultaneous existence of the pair of class and attribute. 
Counterfactuality for a specific visual-linguistic explanation (or region-attribute) can be simply calculated by subtracting classifier outputs corresponding to positive/negative classes of interest.
When the system outputs the explanation, several pairs of [attribute, region] are selected, whose counterfactuality is large for input positive/negative classes.

Maximization (or minimization) of the prediction score with regard to the region is required during the training/inference process, which is computationally intractable in a naive computation. Under the natural assumption that candidate regions are tube-shaped, the maximum (or minimum) value and corresponding region path can be efficiently computed by dynamic programming. We construct the algorithm such that it can be implemented as a layer of a neural network with only standard functions (e.g., max pooling, relu), pre-implemented in most deep learning libraries~\cite{paszke2017automatic, tensorflow2015-whitepaper, chollet2015, jia2014caffe, chainer_learningsys2015} by changing the computation order, which enables combining it easily with Convoluional Neural Networks (CNNs) on GPUs.

The proposed simple and intuitive framework for predicting counterfactuality is justified as the maximization of the lower bound of conditional mutual information, as discussed later, providing an information-theoretical point of view toward it.

Moreover, we assigned additional annotations for existing action-recognition datasets to enable quantitative evaluation for this task, and evaluated our method utilizing the created dataset.

The contributions of this work are as follows:
\begin{itemize}[noitemsep,topsep=0pt]
 \setlength{\itemsep}{0pt}
  \setlength{\parskip}{0pt}
    \item Introduce a novel task, which is generating counterfactual explanations with spatio-temporal region,
    \item Propose a novel method based on simultaneous estimations of visual-linguistic compatibility and discriminativeness of two specific classes,
    \item Propose a novel neural network layer for efficiently solving the dynamic programming problem which appears during training/inference procedures, 
    \item Derive a connection between the proposed method and conditional mutual information maximization, providing better understanding of the proposed model from the information-theoretical viewpoint,
    \item Propose a metric as well as extending datasets for this task, enabling quantitative evaluation,
    \item Demonstrate the effectiveness of the proposed approach by experiment.
\end{itemize}


\section{Related Work}\label{sec:relatedwork}
We divide existing research on visual explanation into two categories, i.e., justification, or introspection. 

Methods for justification are expected to explain as humans do, by projecting input to {\it correct} reason obtained from the outside~\cite{ross2017right}. Methods exploiting textual~\cite{hendricks2016generating} or multimodal~\cite{park2018multimodal, Hendricks_2018_ECCV,Kanehira_2019_CVPR_Learning} supervision belong to this category. True explanations, as by humans, are expected to be generated regardless of the type of model, and evaluation is performed by comparison with ground-truth supervision.

In the latter category, the main goal is to know where the model actually ``looks'' in the input by propagating the prediction to the input space
~\cite{simonyan2013deep, bach2015pixel, zhang2016top, selvaraju2016grad, zhou2016learning, fong2017interpretable, zhou2018interpretable}, or by learning instance-wise importance of elements~\cite{chen2018learning, dabkowski2017real} with an auxiliary model. 
As opposed to the methods of the former category, it is important to show the region where the model focuses for prediction rather than whether the prediction is true for humans. The evaluation is often performed by the investigating before/after output of the model when the element considered to be important is changed. 

Although almost all previous research pursues the reason for positiveness, we attempt to provide the reason for negativeness as well. While our work is categorized to the latter, it also has an aspect of the former; The important region for the model prediction is outputted, while exploiting the linguistic attribute for enhancing interpretability for human.

Beside works of attempting counterfactual explanations by example selection~\cite{goyal2017making, wachter2017counterfactual, Kanehira_2018_CVPR}, ~\cite{Hendricks_2018_ECCV} conducted research similar to the present study, stating an application for grounding visual explanation to counterfactual explanation, where the textual explanations (without the region) are generated by comparing the output of generated explanations for target sample and the nearest sample to it. 
Because their work is in the former category where negativeness cannot be well-defined, no quantitative evaluation was provided.
The main differences between our work and this previous study are:
\begin{itemize}[noitemsep,topsep=0pt]
\item We set generating counterfactual explanations as the main goal, and propose a method which utilizes multimodal information specifically for this task, 
\item Our work belongs to the latter category where negativeness can be well-defined by model output, and therefore, quantitative evaluation is possible,
\item We provide quantitative evaluation with the metric.
\end{itemize}

\begin{figure*}[t]
\begin{center}
\begin{tabular}{c}
\includegraphics[clip, width=0.9\linewidth, height=6.3cm]{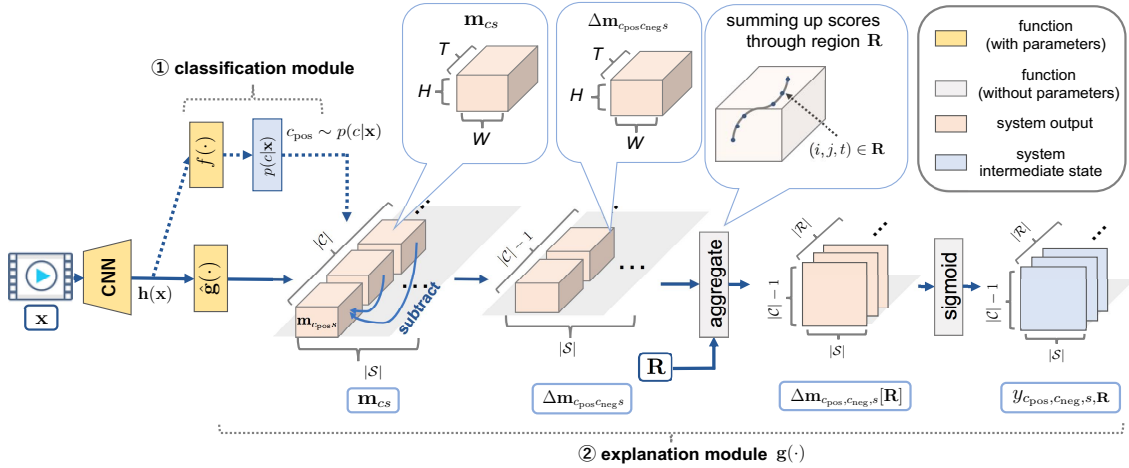}
\end{tabular}
\end{center}
\vspace{-0.8cm}
\caption{Pipeline of the proposed method. Our model holds two modules, the classification and explanation module. The outline of the framework follows two steps: (a) Train a classification model to be explained, (b) Train an auxiliary explanation model in a post-hoc manner by utilizing output and mid-level features of the target classifier after freezing its weights.}
\label{fig:system}
\vspace{-0.4cm}
\end{figure*}

\section{Method}\label{sec:proposed}
We describe the details of our proposed method in this section. The main goal of this work is to build a system that not only classifies a video, but also explains why a given sample is not predicted to one class but another. As stated earlier, we utilize the combination of the spatio-temporal region (tube) and the attribute as the element of explanations. The expected output explanation should have following two properties: 
\begin{enumerate}[noitemsep,topsep=0pt]
\setlength{\itemsep}{0pt}
  \setlength{\parskip}{0pt}
\item[(1)]  Visual explanation is the region which retains high positiveness/negativeness on the model prediction for specific positive/negative classes. 
\item[(2)] Linguistic explanation is compatible to the visual counterpart.
\end{enumerate}

First, we formulate the task addressed in subsection~\ref{subsec:form}, and describe the outline of the framework as well as the actual training/inference process of the explanation model from subsection~\ref{subsec:outline} to~\ref{subsec:train}. In the subsequent subsection~\ref{subsec:mspool} and~\ref{subsec:multiple}, we elucidate the method and its implementation for efficiently solving the optimization problem in the training/inference step. Theoretical background of our method will be discussed in subsection~\ref{subsec:mutalinfo}.

\subsection{Task formulation}\label{subsec:form}
Notations used throughout this paper and formulation of the task we deal with are described in this subsection.

Let ${\mathbf x} \in {\mathbb R}^{W'\times H' \times T' \times 3}$ be the input video where $W', H',T'$ are width, height, and the number of frames of the video, respectively. We denote the class and the attribute by $c \in {\mathcal C}$ and $s \in {\mathcal S}$ , respectively. We assume that attributes are assigned to each sample used for training, and that the assigned set of attributes is represented as ${\mathcal S}({\mathbf x}) \subset {\mathcal S}$. 
${\mathbf R} \subset {\mathcal R}$ denotes the spatio-temporal region used for visual explanation, and its element $[i, j, t, {\rm scale}, {\rm shape}] \in {\mathbf R}$ is the spatio-temporal coordinate, scale, and shape of the element of visual region, respectively. ${\mathcal R}$ is a possible set of ${\mathbf R}$. In this work, we particularly limit ${\mathcal R}$ to the set containing all possible tubes.
In other words, ${\mathbf R}$ contains at most one element corresponding to the time step $t$, and all the elements are spatially and temporally continuous.

We build an explainable model, having the following two functions: 
(a) Classify the video ${\mathbf x}$ to a specific class $c_{\rm pos} \in {\mathcal C}$, 
(b) Explain the reason for specific negative class $c_{\rm neg} \in {\mathcal C}\backslash c_{\rm pos}$ by the combination of attribute $s$ (linguistic) and spatio-temporal tube ${\mathbf R}$ (visual).
Our model predicts several pairs of $(s, {\mathbf R})$ for specific class pair $c_{\rm pos}, c_{\rm neg}$, and simply puts them together as final output.

\subsection{System pipeline}\label{subsec:outline}
We outline the pipeline of the proposed method in Fig.~\ref{fig:system}.
Our model holds two modules, namely, the classification module and the explanation module. The outline of the framework follows two steps:
(a) Train a classification model, which is the target of the explanation,
(b) Train an auxiliary explanation model in a post-hoc manner as in existing research (e.g.,\cite{hendricks2016generating}) by utilizing output and mid-level activation of the target classifier after freezing its weights to prevent change in output. 
Specifically, we explicitly represent feature extraction parts in the pre-trained classification network as 
\begin{eqnarray}\label{eq:classifier}
p(c|{\mathbf x}) = f({\mathbf h}({\mathbf x})),\ {\mathbf h}({\mathbf x}) \in {\mathbb R}^{W \times H \times T \times d},
\end{eqnarray}
where $W, H, T$ indicate width, height, and the number of frames in the mid-level feature, respectively. The $d$-dimensional feature vector corresponding to each physical coordinate is denoted by ${\mathbf h}({\mathbf R})[i, j, t] \in {\mathbb R}^d$.

We introduce an auxiliary model ${\mathbf g}$, which is responsible for the explanation. It predicts {\it conterfactuality}, which measures (1) the positiveness/negativeness of the region ${\mathbf R}$ on the model prediction $p(c|{\mathbf x})$ for a specific pair of $c_{\rm pos}, c_{\rm neg}$, and (2) the compatibility of linguistic explanation $s$ to the visual counterpart ${\mathbf R}$. By fixing the parameter of the feature extraction part ${\mathbf h}(\cdot)$, we obtain
\begin{eqnarray}
y_{c_{\rm pos}, c_{\rm neg}, s, {\mathbf R}}  = {\mathbf g}({\mathbf h}({\mathbf x})) \in [0, 1 ]^{(|{\mathcal C}| - 1)\times |{\mathcal S}|\times |{\mathcal R}|} 
\end{eqnarray}
which holds a counterfactuality score corresponding to one combination of $c_{\rm pos}, c_{\rm neg}, s, {\mathbf R}$ in each dimension.
Positive class $c_{\rm pos}$ is sampled from $p(c|{\mathbf x})$ during training, and $c_{\rm pos}= \argmax_{c} p(c|{\mathbf x})$ is applied in the inference step. Any remaining class $c_{\rm neg}\in {\mathcal C}\backslash c_{\rm pos}$ is regarded as negative.

We consider the element of ${\mathbf R}$ in the space of ${\mathbf h}({\mathbf x})$. In other words, the coordinate $(i, j, t)$ of ${\mathbf R}$ corresponds to that of ${\mathbf h}({\mathbf x})$. The shape of ${\mathbf R}$ is fixed to $[W/W', H/H']$ for the sake of simplicity. The extension to multiple scales and the aspect ratio will be discussed in subsection~\ref{subsec:multiple}.

\subsection{Predicting counterfactuality}\label{subsec:train}
Our explanation model predicts counterfactuality, that is, (1) how much the region ${\mathbf R}$ retains high positiveness/negativeness on the $p(c|{\mathbf x})$ for $c_{\rm pos}, c_{\rm neg}$, and (2) how much $s$ is compatible to ${\mathbf R}$. The counterfactuality score is predicted in the following steps.

A target sample ${\mathbf x}$ is inputted to obtain the mid-level representation ${\mathbf h}({\mathbf x})$ and the conditional probability $p(c|{\mathbf x})$. 
The explanation model holds classifiers ${\hat {\mathbf g}_{cs}}$ for each pair of $(c, s)$. These classifiers are applied to each element feature of ${\mathbf h}({\mathbf x})$, and predict simultaneous existence of $(c, s)$ as 
\begin{eqnarray}\label{eq:inner-func}
{\mathbf m}_{cs} = {\hat {\mathbf g}_{cs}}({\mathbf h}({\mathbf x}))\in {\mathbb R}^{W \times H \times T}, 
\end{eqnarray}
is obtained for each pair of $(c, s)$.
For simplicity, we utilize a linear function that preserves geometrical informatiton, that is, the convolutional layer as classifier ${\hat {\mathbf g}}_{cs}$.

To measure how likely the region element is considered to be $c_{\rm pos}$, not $c_{\rm neg}$, with linguistic explanation $s$, we element-wise subtract the value of (\ref{eq:inner-func}) for all $c_{\rm neg} \in {\mathcal C} \backslash \{c_{\rm pos}\}$ and $s$ as 
\begin{eqnarray}\label{eq:deltam}
\Delta {\mathbf m}_{c_{\rm pos}, c_{\rm neg}, s} =  {\mathbf m}_{c_{\rm pos}s} - {\mathbf m}_{c_{\rm neg}s}
\end{eqnarray}

To obtain the score for the region ${\mathbf R}$, we define the procedure of  aggregating scalar values through ${\mathbf R}$ in the 3 dimensional tensor $\Delta {\mathbf m}$ as 
\begin{eqnarray}
\Delta {\mathbf m}_{c_{\rm pos}, c_{\rm neg}, s}[{\mathbf R}]
= \sum_{(i, j, t) \in {\mathbf R}} \Delta {\mathbf m}_{c_{\rm pos}, c_{\rm neg}, s}[i, j, t].
\end{eqnarray}
Please note $\Delta {\mathbf m}_{c_{\rm pos}, c_{\rm neg}, s}[{\mathbf R}] \in {\mathbb R}$.
By applying a sigmoid activation function to the output, we obtain counterfactuality $y_{c_{\rm pos}, c_{\rm neg}, s, {\mathbf R}} \in [0, 1 ]^{(|{\mathcal C}| - 1)\times |{\mathcal S}|}$ as 
\begin{eqnarray}\label{eq:out}
y_{c_{\rm pos}, c_{\rm neg}, s, {\mathbf R}} = \sigma(\Delta {\mathbf m}_{c_{\rm pos}, c_{\rm neg}, s}[{\mathbf R}])
\end{eqnarray}
where $\sigma(a) = \frac{1}{1 + {\rm exp}(-a)}$.

\subsection{Training and inference}\label{subsec:train}
We illustrate the loss function optimized in the training step and the procedure in the inference step. 

\textbf{Loss function}:
The supervised information ``A sample is classified to category $c_{\rm pos}$.'' can be translated to ``A sample is not classified to any category $c_{\rm neg} \in {\mathcal C}\backslash c_{\rm pos}$.'' By utilizing this, the model is trained to enlarge the counterfactuality score corresponding to class pairs ${c_{\rm pos}, {c_{\rm neg}}}$ and attributes $s \in {\mathcal S}({\mathbf x})$. The output obtained after sigmoid activation in (\ref{eq:out}) can be interpreted as a probability, and its negative log likelihood is minimized. 
Because computing output $y_{c_{\rm pos}, c_{\rm neg}, s, {\mathbf R}}$ for all pairs of $c_{\rm neg} \in {\mathcal C}\backslash c_{\rm pos}$, $s\in{\mathcal S}$ and ${\mathbf R}\in {\mathcal R}$ is not feasible, only ${\mathbf R}$ maximizing the loss is utilized for each pair of $c_{\rm neg}, s$ while training. Formally, for a given ${\mathbf x}$ and $c_{\rm pos}$, the loss 
\begin{eqnarray}\label{eq:train}
{\ell}({\mathbf x}, c_{\rm pos}) = \frac{1}{|{\mathcal S({\mathbf x})}|}\sum_{s\in {\mathcal S({\mathbf x})}}\sum_{c_{\rm neg}\in {\mathcal C}\backslash c_{\rm pos}}-{\rm log}\ {\hat y_{c_{\rm pos}, c_{\rm neg}, s}}\nonumber \\
{\rm where}\ \ {\hat y_{c_{\rm pos}, c_{\rm neg}, s}} = \min_{\mathbf R}\ y_{c_{\rm pos}, c_{\rm neg}, s, {\mathbf R}}
\end{eqnarray}
is calculated.
As stated in the next subsection~\ref{subsec:mspool}, 
${\hat y_{c_{\rm pos}, c_{\rm neg}, s}}$ can be efficiently computed by dynamic programming under the condition where ${\mathcal R}$, a possible set of ${\mathbf R}$, is limited to the set of all spatio-temporal tubes.

The overall loss function to be optimized is obtained by taking the expectation of (\ref{eq:train}) over ${\mathbf x}$ and $c_{\rm pos}$ as
\begin{eqnarray}\label{eq:loss}
{\mathcal L} = {\mathbb E}_{p(\mathbf{x})p(c_{\rm pos}|\mathbf{x})}
\left[ {\ell}({\mathbf x}, c_{\rm pos})\right]. 
\end{eqnarray}

$p(\mathbf x)$ indicates the true sample distribution, and $p(c_{\rm pos} | {{\mathbf x})}$ is the pre-trained network in (\ref{eq:classifier}).
Empirically, the expectation over ${\mathbf x}$ is calculated by summing up all training $N$ samples, and that over $c_{\rm pos}$ is achieved by sampling from the conditional distribution $p(c_{\rm pos} | {{\mathbf x})}$ given ${\mathbf x}$.

\textbf{Inference}: During inference, provided with positive and negative class $c_{\rm pos}, c_{\rm neg}$ as well as input ${\mathbf x}$, pairs of the attribute $s$ and the region ${\mathbf R}$ are outputted whose score is the largest. Formally, 
\begin{eqnarray}\label{eq:inference}
s^{\star}, {\mathbf R}^{\star} = \argmax_{s, {\mathbf R}}\ y_{c_{\rm pos}, c_{\rm neg}, s, {\mathbf R}}
\end{eqnarray}
is calculated as the element of explanation.
For computing $k$ multiple outputs, we compute $\max_{\mathbf R}\ y_{c_{\rm pos}, c_{\rm neg}, s, {\mathbf R}}$ for all $s$ and pick $k$ pairs whose scores are the largest. Minimization for ${\mathbf R}$ is also efficiently calculated as is the case in the training step.

\begin{figure}[t]
\begin{center}
\begin{tabular}{c}
\includegraphics[clip, width=\linewidth, height=4.5cm]{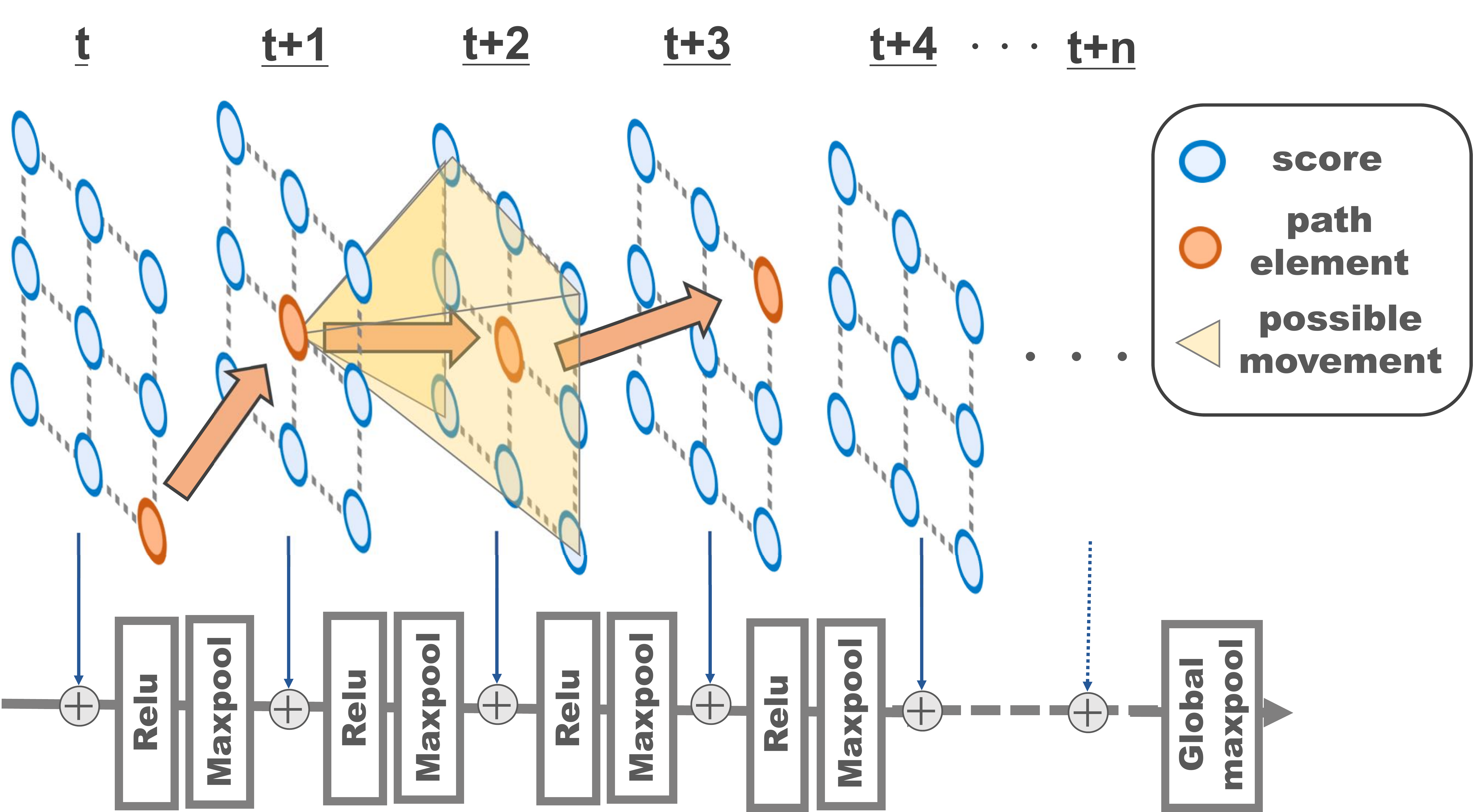}
\end{tabular}
\end{center}
\vspace{-0.5cm}
\caption{The illustration of maximum subpath pooling. Finding the subpath in the 3d tensor whose summation is maximum can be implemented by sequentially applying the elementwise sum, relu, and 2d max pooling in the time direction, following global 2d max pooling.}
\vspace{-0.5cm}
\label{fig:motivation}
\end{figure}

\subsection{Maximum subpath pooling}\label{subsec:mspool}
We describe in detail the maximization (minimization) problem for ${\mathbf R}$ appearing in (\ref{eq:train}) and (\ref{eq:inference}) in this subsection.
We limit ${\mathcal R}$, a possible set of ${\mathbf R}$, to the set containing all possible tubes. A tube can be expressed as a path in the 3d tensor, starting at one spatial coordinate $(i, j)$ in time $t$, and move to the neighbor $\{(i+l, j+m)\ |\ -k\le l, m \le k\}$ in time $t+1$ where $k$ controls how much movement of the spatial coordinate $(i, j)$ is allowed when the time changes from $t$ to $t+1$. 
The path can start and end at any time. The ${\mathbf R}$ consists of coordinates $(i, j, t)$ satisfying the path condition. 

With this limitation, the maximization problem with regard to ${\mathbf R}$ can be cast as an extension of finding a sub-array whose summation is maximum, and it can be efficiently solved by the algorithm proposed in ~\cite{tran2014video} (shown in the supplementary material), which is an extension of the Kadane's algorithm~\cite{bentley1984programming}.
Although~\cite{tran2014video} utilized it only for the inference, we need to train the parameters, especially by back-propagation on GPUs to combine CNNs. To realize this, we construct the algorithm such that it can be implemented as a layer of a neural network with only standard functions pre-implemented in most deep learning libraries~\cite{paszke2017automatic, tensorflow2015-whitepaper, chollet2015, jia2014caffe, chainer_learningsys2015}. Interestingly, as shown in Fig.~\ref{fig:motivation}, the same result can be achieved by sequentially applying relu, 2d maxpooling, and element-wise summation in the time direction followed by global 2d maxpooling. 
The kernel size of maxpooling is a hyper-parameter corresponding to $k$ mentioned above. We fixed it to $3\times3$, which means that $k=1$. The computational cost of this algorithm is $O(WHT)$, which can be solved by a single forward path, since the iteration for $W$ and $H$ can be parallelized on GPU without significant overhead.
In the case of minimization of the objective, the same algorithm can be applied just by inverting sign of input and output.

To acquire the path ${\mathbf R}^{\star}$ whose summation is maximum, we simply need to calculate the partial derivative of the maximum value with regards to each input. Because

$\frac{\partial \Delta {\mathbf m}_{c_{\rm pos}, c_{\rm neg}, s}[{\mathbf R}^{\star}]}{\partial  \Delta {\mathbf m}_{c_{\rm pos}, c_{\rm neg}, s}}[i, j, t] = \begin{cases}
    1 & ((i, j, t) \in {\mathbf R}^{\star}) \\
    0 & ({\rm otherwise})
  \end{cases}
$ \\
we can obtain the path corresponding to the maximum by extracting the element whose derivative is equal to 1. Implementation is likewise easy for the library, which has the function of automatic differentiation.

This procedure can be interpreted as a kind of pooling. 
To observe this, we denote the aggregated feature after applying sum pooling to mid-level local features throughout the region ${\mathbf R}^{\star}$ by ${\rm pool}({\mathbf h}({\mathbf x}), {\mathbf R}^{\star})$, and redefine $\Delta{\mathbf w}={\mathbf w}_{c_{\rm pos}s}-{\mathbf w}_{c_{\rm neg}s}$, where ${\mathbf w}_{cs}$ is the parameter of the convolutional layer ${\hat {\mathbf g}}_{cs}$. The summation of the score inside the sigmoid function in (\ref{eq:inference}) can be written as 
\begin{eqnarray}
\max_{\mathbf R} \Delta {\mathbf m}_{c_{\rm pos}, c_{\rm neg}, s}[{\mathbf R}]
=\sum_{(i, j, t)\in{\mathbf R}^{\star}} \Delta{\mathbf w}^{\top} {\mathbf h}({\mathbf x})[i, j, t]  \nonumber \\
=\Delta{\mathbf w}^{\top} \sum_{(i, j, t)\in{\mathbf R}^{\star}} {\mathbf h}({\mathbf x})[i, j, t] 
=\Delta{\mathbf w}^{\top} {\rm pool}({\mathbf h}({\mathbf x}), {\mathbf R}^{\star}) \nonumber
\end{eqnarray}
We refer to the sequence of this process as the maximum subpath pooling layer.

\begin{figure}[t!]
  \includegraphics[width=\linewidth]{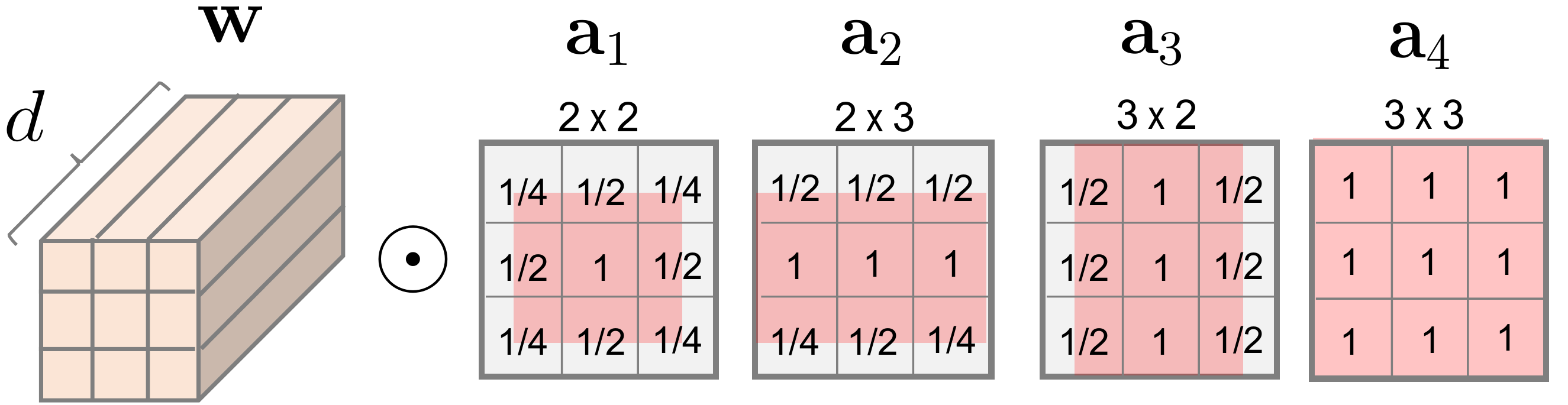}
  \vspace{-0.4cm}
  \caption{The spatial weights multiplied with the parameters of the convolutional layer. Each element of the weight has a value proportional to the overlap to the outputted shape (in red). These values are normalized such that the summation equals 1.}
  \vspace{-0.6cm}
\label{fig:mask}
\end{figure} 

\subsection{Multiple scales and aspect ratio}\label{subsec:multiple}
So far, we only considered the situation where the scale and aspect ratio of ${\mathbf R}$ is fixed to $[W/W', H/H']$, $1:1$. 
We modify the algorithm to treat different scales and shapes of the region. 
As described in~\ref{subsec:mspool},
the input of the optimization algorithm $\Delta {\mathbf m}_{c_{\rm pos},c_{\neg},s}$ (defined in (\ref{eq:deltam})) is a 3d tensor, which corresponds to each physical coordinate in the region. We expand the input from 3d to 5d by expanding ${\mathbf m}_{c,s}$ (defined in (\ref{eq:inner-func})),
taking scales and shapes into consideration. 
To obtain the 5d tensor, we prepare multiple parameters ${\mathbf w}$ of the convolutional layer ${\hat {\mathbf g}_{cs}}$ (\ref{eq:inner-func}) for each scale and shape. (For simplicity, the subscripts $c, s$ of ${\mathbf w}$ will be omitted below). 
partial
To treat different scales, we extract  mid-level representations from different layers of the target classifier to construct ${\mathbf h}({\mathbf x})$. 
After convolution is applied separately, they are appended for the scale-dimension.

When considering different shapes at each scale, we prepare different parameters by multiplying the significance corresponding to the shape of a region. Formally, ${\mathbf w}_{i}={\mathbf w} \odot {\mathbf a}_i$ is computed, where ${\mathbf a}_i$ has the same window size as ${\mathbf w}$ and consists of the importance weight for each position of parameter, which is determined by the overlap ratio between the region and each local element. They are normalized to satisfy $|{\mathbf a}_i| = 1$. Different ${\mathbf w}_i$ are applied separately and output is appended to the shape-dimension of the tensor. Concretely, we compute the scores corresponding to four kinds of shapes $(2\times 2, 2\times 3, 3\times 2, 3 \times 3 )$ from $3 \times 3$ convolution for each scale as in Fig.~\ref{fig:mask}.
The optimization problem can be solved by applying the same algorithm described in subsection~\ref{subsec:mspool} to the obtained 5d tensor.

\subsection{Theoretical background}\label{subsec:mutalinfo}
To demonstrate that linguistic explanation $s$ for region ${\mathbf R}$ is strongly dependent on the output of the target classifier $c$ by minimizing the loss function proposed above, 
we reveal the relationship between the loss function and conditional mutual information.
Conditional mutual information of the distribution parameterized by our model is denoted by ${\rm MI}(c, s| {\mathbf x}, {\mathbf R})$  and can be bounded as 
\begin{eqnarray}\label{eq:mi}
{\rm MI}(c, s| {\mathbf x}, {\mathbf R}) &=& \mathbb{E}_{p(c, s, {\mathbf x}, {\mathbf R})}\left[ \rm{log}\ \frac{{\hat p}(c, s | {\mathbf x}, {\mathbf R})}{{\hat p}(c| {{\mathbf x}, \mathbf R}){\hat p}(s | {\mathbf x}, {\mathbf R})}\right] \nonumber \\
&\ge& \mathbb{E}_{p(c, s, {\mathbf x}, {\mathbf R})}\left[{\rm log}\ {\hat p}(c | s, {\mathbf x}, {\mathbf R})\right]\label{eq:a}
\end{eqnarray}
(\ref{eq:a}) is derived from ${\rm H}(c|{\mathbf x}, {\mathbf R})\ge 0$ and ${\rm KL}[p | q]\ge 0$ for any distribution $p, q$ where ${\rm H}(\cdot)$ and ${\rm KL}[\cdot|\cdot]$ indicate the entropy and KL divergence, respectively. In our case, we parameterize the joint distribution as
\begin{table}[t!]
\small 
  \centering
   \begin{tabular}{|c||c|c|c|c|} \hline 
    dataset & video & class & attribute & bbox  \\ \hline\hline 
    Olympic & 783 & 16 &  39 & 949  \\ \hline 
    UCF101-24 & 3204 & 24 & 40 & 6128  \\ \hline 
   \end{tabular}
   \vspace{-0.3cm}
    \caption{statistics of dataset used in the experiment}
      \vspace{-0.9cm}
  \label{tb:statistics}
  
\end{table}
\begin{eqnarray}\label{eq:score}
{\hat p}(c, s | {\mathbf x}, {\mathbf R}) &=& \frac{\rm{exp}({\mathbf m}_{cs}[{\mathbf R}])}{\sum_{c'\in C, s'\in  S}\rm{exp}({\mathbf m}_{c's'}[{\mathbf R}])}
\end{eqnarray}
(\ref{eq:a}) is further bounded as
\begin{flalign}
(\ref{eq:a}) 
\ge& \mathbb{E}_{p(c, s, {\mathbf x})}\left[ \min_{\mathbf R}\ \rm{log}\frac{\rm{exp}({\mathbf m}_{cs}[{\mathbf R}])}
{\sum_{c's'}\rm{exp}({\mathbf m}_{c's'}[{\mathbf R}])} \right]  \label{eq:c}&&\\
\ge& \mathbb{E}_{p(c_{\rm pos}, s, {\mathbf x})}\!\left[\! \sum_{c_{\rm neg}\in {\mathcal C}\backslash c_{\rm pos}} \!\!\!\!\!\!\min_{\mathbf R} \sigma(\Delta{\mathbf m}_{c_{\rm pos}, c_{\rm neg}, s}[{\mathbf R}]) \! \right] \label{eq:d} &&\\
=& \mathbb{E}_{p(c_{\rm pos}, s, {\mathbf x})}\left[ \sum_{c_{\rm neg}\in {\mathcal C}\backslash c_{\rm pos}} {\hat y_{c_{\rm pos}, c_{\rm neg}, s}} \right] &&
\end{flalign}
$\sigma(\cdot)$ is the sigmoid function and (\ref{eq:c}) is derived from the fact $\mathbb{E}_{a}[{\rm f}(a)] \ge \min {\rm f}(a)$.
On the bound (\ref{eq:d}), we utilize the relationship $(1+\sum_{i} a_i) \le \prod_{i} (1 + a_i)$~\cite{aueb2016one}.

Finally, by decomposing as $p(s, c| {\mathbf x}) = p(s | c, {\mathbf x})p(c | {\mathbf x})$ and setting $p(c | {\mathbf x})$ as the target classifier and 
$p(s | c, {\mathbf x}) = [s \in {\mathcal S}({\mathbf x})] / |{\mathcal S}({\mathbf x})|$ following the inversion of sign, (\ref{eq:loss}) is obtained. The minimization of the loss function can be justified as the maximization of the lower bound of conditional mutual information.
It may be beneficial to investigate the relationship with other methods proposed for the explanation task, such as~\cite{chen2018learning}, based on mutual information maximization.

\section{Experiment}\label{sec:experiment}
We describe experiments to demonstrate the effectiveness of proposed method, in particular the explanation module, which is the main proposal for this task. After the details of experimental settings including datasets and metrics used for quantitative evaluation are described
in subsection~\ref{subsec:setting}, the obtained results are discussed in subsection~\ref{subec:neg}.

\subsection{Setting}\label{subsec:setting}
Given an input video ${\mathbf x}$, and a pair of positive/negative class $c_{\rm pos}, c_{\rm neg}$, the explanation module outputs several pairs of attribute/region $\{(s_i, {\mathbf R}_i)\}_{i=1}^{k}$. We separately evaluate each pair of output.

\begin{figure}[t!]
\centering
\begin{minipage}{.45\textwidth}
  \centering
  \includegraphics[width=.9\linewidth]{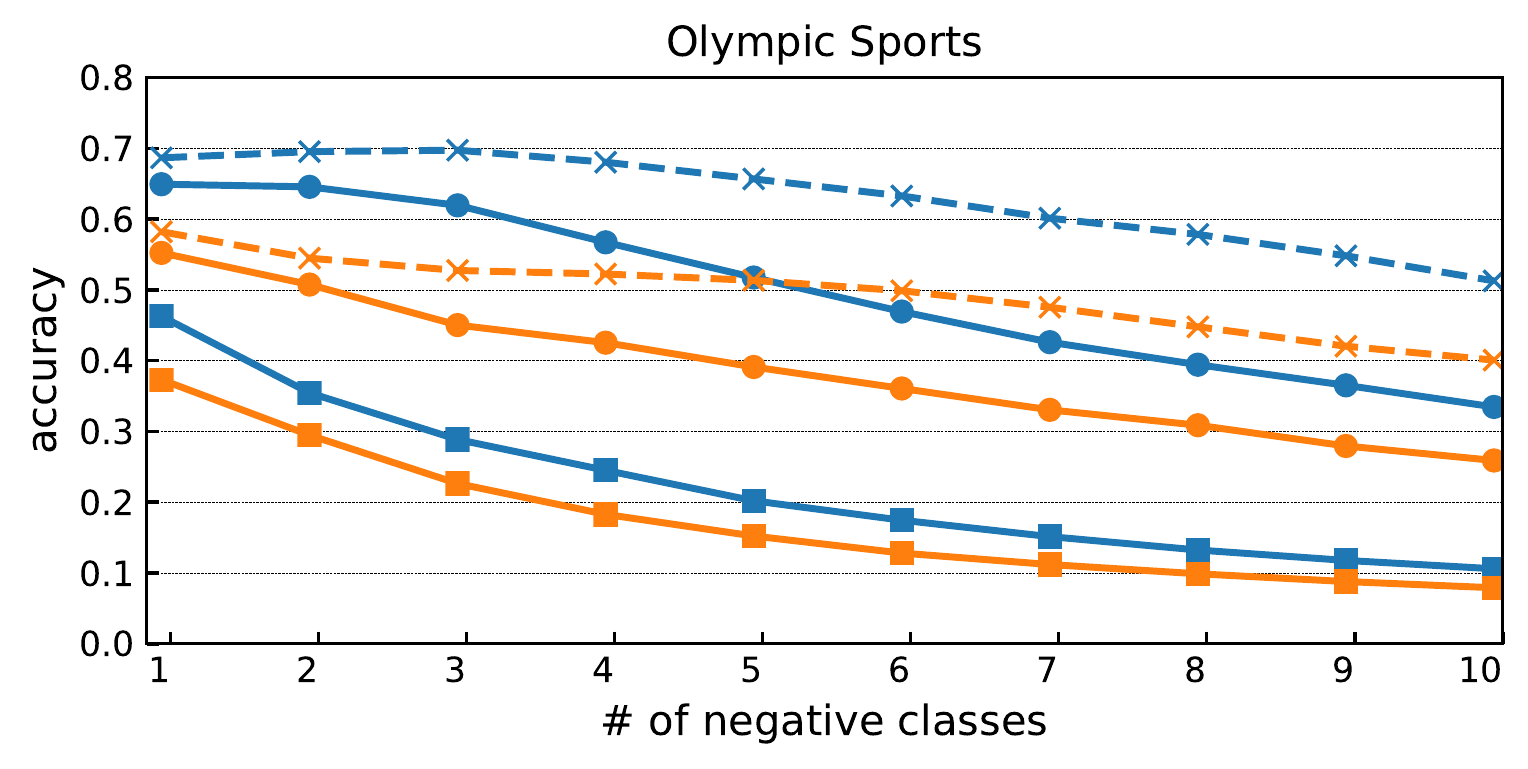}
\end{minipage}\\ \vspace{-0.1cm}

\begin{minipage}{.45\textwidth}
  \centering
  \includegraphics[width=.9\linewidth]{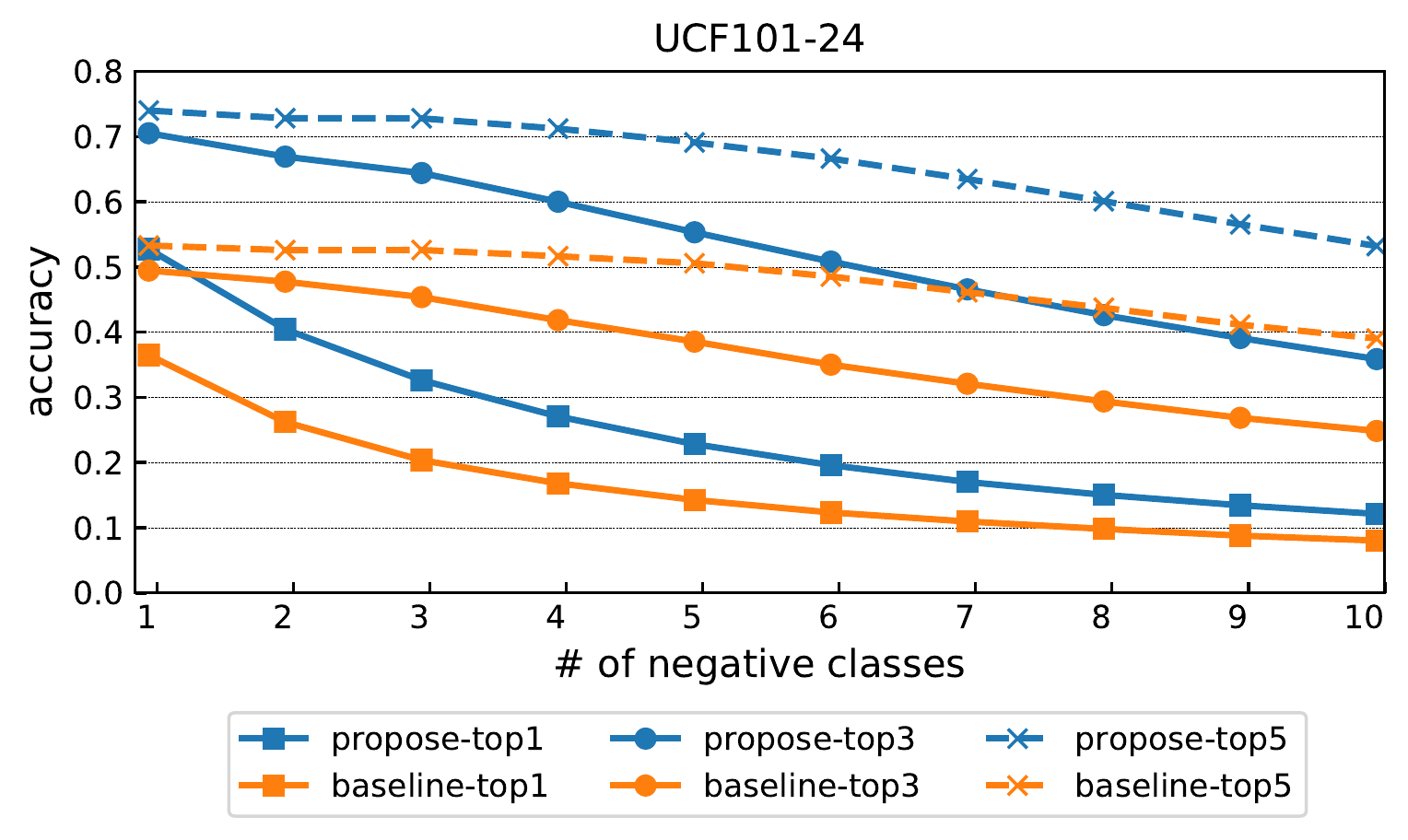}
\end{minipage}
\vspace{-0.3cm}
\caption{The negative class accuracy on the Olympic Sports dataset (above) and the UCF101-24 dataset (below). The y-axis depicts the mean accuracy and the x-axis denotes the number of negative classes used for averaging, whose prediction value is maximum.}
\label{fig:negacc}
\vspace{-0.70cm}
\end{figure}

\textbf{Dataset}:
Two existing video datasets for action recognition: Olympic Sports~\cite{niebles2010modeling} and UCF101-24 categories~\cite{THUMOS14} were used in the experiments. 
The Olympic Sports dataset consists of 16 categories for sports action. The UCF101-24 categories is a subset of the UCF101 dataset~\cite{soomro2012ucf101} extracting 24 out of 101 general action classes. We utilized the original train/test split provided by the datasets.
We additionally assigned these datasets with Amazon Mechanical Tutk (AMT) to make the evaluation possible as follows: 
(a) Assign a set of attributes to all videos in the dataset,
(b) Assign a bounding box of assigned attributes for samples in the test split.
Statistics of the dataset are shown in Table.~\ref{tb:statistics}, and a few examples of the annotations are provided in the supplementary material.

\textbf{Metric}:
As stated earlier, the method for this task is expected to satisfy the following two requirements:
\begin{enumerate}[noitemsep,topsep=0pt]
\setlength{\itemsep}{0pt}
  \setlength{\parskip}{0pt}
\item[(1)] Visual explanation is the region which retains high positiveness/negativeness on the model prediction for specific positive/negative classes,
\item[(2)] Linguistic explanation is compatible to the visual counterpart,
\end{enumerate}
and methods are evaluated based on them.
To make the quantitative evaluation possible, we propose two metrics, both of which are based on the accuracy.

As for (1), we need to evaluate whether the obtained region is truly an explanation of the reason for  
``the target classifier predicts a sample to not $c_{\rm neg}$ but $c_{\rm pos}$''. More specifically, we would like to confirm whether the region explains the specific negative class $c_{\rm neg}$, not other negatives ${\hat c_{\rm neg}} \in {\mathcal C}\backslash \{c_{\rm pos}, c_{\rm neg} \}$. To evaluate this quantitatively, we investigate how the output of the target classifier changes corresponding to $c_{\rm neg}$ when region ${\mathbf R}$ is removed from the input.
A mask ${\mathbf z} = \{0, 1\}^{W'\times H' \times T'\times 3}$ is prepared, which takes the value of 0 if the corresponding pixel's location is contained to ${\mathbf R}$ restored on the input space, otherwise it takes the value of 1. Applying again the masked sample to the target classifier, the difference from the original output 
$f({\mathbf h}({\mathbf x} \odot {\mathbf z})) - f({\mathbf h}({\mathbf x}))$ is calculated for all negative classes where $\odot$ denotes the Hadamard product.
We calculate the accuracy, i.e., we pick the largest values out of the obtained difference scores, and examine if $c_{\rm neg}$ exists within this set. We refer to this metric as negative class accuracy. 

As for (2), we assess how the region ${\mathbf R}$ makes the concept $s$ identifiable by humans for each output pair $(s, {\mathbf R})$. To quantify this, we exploit the bounding boxes assigned for each attribute in the test set, and compute the accuracy as follows.
IoU (intersection over union) is calculated between given ${\mathbf R}$ and all bounding boxes ${\mathbf R}'$, which corresponds to attribute $s'$. We measure the accuracy by selecting the attribute $s'$ with the largest IoU score, and checking its consistency with $s$, which is the counterpart of ${\mathbf R}$. This metric is referred to as concept accuracy in the following parts.

\begin{table}[t!]
\small 
  \centering
   \begin{tabular}{|c||c|c|} \hline 
    method & Olympic & UCF101-24  \\ \hline\hline 
    baseline & 0.76 & 0.68    \\ \hline 
    propose & 0.89 & 0.88    \\ \hline 
   \end{tabular}
   \vspace{-0.3cm}
    \caption{The ratio of the probability $p(c_{\rm pos}|{\mathbf x})$ for the positive class $c_{\rm pos}$ decreasing after the region is masked out.}
  \label{tab:posacc}
   \vspace{-0.1cm}
\end{table}

\begin{table}[t!]
\small 
    \centering
        \label{multiprogram}
        \begin{tabular}{|c||c|c|c|c|c|c|}
            \hline
        \  & \multicolumn{3}{|c|}{Olympic Sports}& \multicolumn{3}{|c|}{UCF101-24}\\
            \hline 
            method & top1 & top3 & top5 & top1 & top3 & top5  \\
            \hline
            \hline
            baeline & 0.02 & 0.07 & 0.12 & 0.02 & 0.07 & 0.13 \\
            \hline
            propose & 0.13 & 0.38 & 0.59 & 0.14 & 0.41 & 0.65 \\
            \hline
        \end{tabular}
            \vspace{-0.3cm}
\caption{The concept accuracy on the Olympic Sports dataset and the UCF101-24 dataset.}
    \label{tb:cocept}
         \vspace{-0.5cm}
    \end{table}

\textbf{Detailed settings}
In the classification module, the output of convolutional layers was used as ${\mathbf h}(\cdot)$. Fc layers following convolutional layers were considered as $f(\cdot)$. 

Specifically, we dealt with C3D-resnet~\cite{tran2017convnet, hara3dcnns} as the target classifier in the experiments, which is based on the spatio-temporal convolution~\cite{tran2015learning} with 
residual architecture~\cite{he2016deep}. 
Our network for classification consists of nine convolutional layers and one fully connected layer accepting a $112\times 112\times 16$ size input. Relu activation was applied to all layers, except the final fc layer.
We selected the outputs of the last and the 2nd to last convolutional layers to construct ${\mathbf h}(\cdot)$. 
Moreover, we replaced 3d max pooling to 2d max pooling to guarantee $T = T'$ for all activations.
The target classifier was trained with SGD, where learning rate, momentum, weight decay, and batch size, were set to 0.1, 0.9, 1e-3, and 64 respectively. To train the model in Olympic Sports, we pre-trained it with the UCF101-24. 

For the training of the explanation module, all the settings (e.g., learning rate) were set to the same as in the training of classification module, except that the batch size was 30. We decomposed the weight of the convolutional layer ${\mathbf w}_{cs}$ corresponding to the pair of $(c, s)$ to ${\mathbf w}_{cs}={\mathbf w}_{c}\odot {\mathbf w}_{s} +{\mathbf w}_{c}+{\mathbf w}_{s}$
to reduce the number of parameters where ${\mathbf w}_{s}$ and ${\mathbf w}_{c}$ are the parameter shared by the same attribute and class respectively.

\subsection{Identifiability of negative class}\label{subec:neg}
To assess whether the obtained region ${\mathbf R}$ is truly an explanation of the reason for a specific negative class, we compared the negative class accuracy with a baseline.
Because there is no previous work for this task, we employed a simple baseline. For the UCF101-24 dataset, we exploited the bounding box for the action detection provided in~\cite{singh2016online}, which provides an upper limit of the performance for the action detection task.
As bounding boxes are not provided for the Olympic Sports dataset, we cropped the center ($32 \times 32$) from all frames. This forms a simple but strong baseline because the instance containing category information usually appears in the center of the frame in this dataset.

\begin{table}[t!]
\small 
    \centering
        \begin{tabular}{|c||c|c|c|c|c|c|}
            \hline
        \  & \multicolumn{3}{|c|}{Olympic Sports}& \multicolumn{3}{|c|}{UCF101-24}\\
            \hline 
            method & fc1 & fc2 & fc3 & fc1 & fc2 & fc3  \\
            \hline
            \hline
            baeline & 0.45 & 0.44 & 0.46 & 0.45 & 0.43 & 0.44 \\
            \hline
            propose & 0.62 & 0.64 & 0.59 & 0.64 & 0.62 & 0.61 \\
            \hline
        \end{tabular}
         \vspace{-0.3cm}
             \caption{The top3 negative class accuracy on the Olympic Sports dataset and the UCF101-24 dataset averaged over 3 negative classes whose prediction probability is the largest, by changing the number of fully-connected layers.}
    \label{tb:fc}
    \vspace{-0.5cm}
    \end{table}

The results of the negative class accuracy for the Olympic Sports and the UCF101-24 datasets are shown in Fig.~\ref{fig:negacc}. The accuracy averaged over negative classes having largest $p(c | {\mathbf x})$ is calculated and the x-axis of the figures depicts the number of used negative classes. 
Our method performed consistently better than the baseline on both datasets, demonstrating the generalization of identifiability of negativeness in unseen samples. The gap between the accuracy of our method and that of baseline is decreased when negative classes having small $p(c | {\mathbf x})$ are included. We conjecture the reason is that such a {\it easy negative} class, which is highly dissimilar to the positive class, does not have common patterns to identify them. For example, for positive class `pole vault', the negative class `high jump' is considered to be more similar than `bowling' for the classifier, and detecting the negativeness for such a {\it hard negative} `pole vault' is relatively easy (e.g. the region of `pole'), although detecting it is difficult for the {\it easy negative} class.
We believe the low-accuracy for such an {\it easy negative} class does not have significant impact because in real applications, 
we may be interested in the reason for the negativeness of the {\it hard negative} class, which is difficult to discriminate.
In addition, we also report the ratio of the probability $p(c_{\rm pos}|{\mathbf x})$ for the positive class $c_{\rm pos}$ decreasing after the region is masked out in Table.~\ref{tab:posacc}. From these results, we claim that our method can find both better of negativeness/positiveness on the negative/positive classes.

\subsection{Identifiability of concept}
To evaluate whether the obtained region makes the concept (linguistic explanation) identifiable by human, we measured concept accuracy described above for the case where category prediction is correct. The same baseline was applied for region selection as in the previous subsection~\ref{subec:neg}, and the attribute is randomly selected. 
Results are shown in Table~\ref{tb:cocept}. In both datasets, our method is consistently better than the baseline. Finding the region by which a specific concept can be identified is the significantly challenging task, where methods need to identify small objects or atomic-actions~\cite{gu2017ava}. Although it is conceivable that there is still room for improvement of this metric, we believe that our simple method can serve as a base for future work regarding this novel task. 

\subsection{Influence of the complexity of classifier}
To investigate the influence of the complexity of the classifier module on the generalization ability of the explanation module, we measured the negative class accuracy by changing the number of fc-layers of $f(c | {\mathbf x})$ in 1 $\sim$ 3. The other settings remain the same as those in subsection~\ref{subec:neg}.
The top3 accuracy averaged on 3 negative classes is shown in Table.~\ref{tb:fc} (Other results are shown in supplementary material). In both datasets, the gap between the baseline and the proposed method is consistent regardless of the number of fc-layers, demonstrating the robustness of the proposed method to the complexity of classifiers to be explained.

\begin{figure}[t!]
  \centering
  \includegraphics[width=0.95\linewidth]{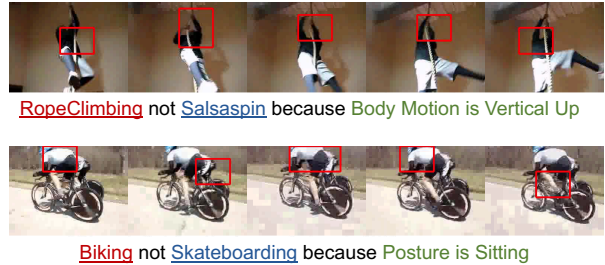}
  \vspace{-0.35cm}
  \caption{Example output from our system for the samples of the UCF101-24 dataset.}
  \label{fig:examples}
    \vspace{-0.6cm}
\end{figure}

\subsection{Output examples}
We show a few output examples in Fig.~\ref{fig:examples} for the sample videos of the UCF101-24 dataset. As observed in the figures, our model is considered to appropriately localize the area compatible with the linguistic explanation. Other examples are shown in the supplementary material.

\section{Conclusion}\label{sec:conclusion}
In this work, we particularly focused on building a model that not only categorizes a sample, but also generates an explanation with linguistic and region. To this end, we proposed a novel algorithm to predict {\it counterfactuality}, while identifying the important region for the linguistic explanation. Furthermore, we demonstrated the effectiveness of the approach on two existing datasets extended in this work.

\section{Acknowledgement}
This work was partially supported by JST CREST Grant Number JPMJCR1403, Japan, and partially supported by the Ministry of Education, Culture, Sports, Science and Technology (MEXT) as "Seminal Issue on Post-K Computer." Authors would like to thank Kosuke Arase, Mikihiro Tanaka, Yusuke Mukuta for helpful discussions.

{\small
\bibliographystyle{ieee}
\bibliography{egbib}
}

\end{document}


\title{Multimodal Explanations by Predicting Counterfactuality in Videos \\ (supplementary material)}

\author{First Author\\
Institution1\\
Institution1 address\\
{\tt\small firstauthor@i1.org}
\and
Second Author\\
Institution2\\
First line of institution2 address\\
{\tt\small secondauthor@i2.org}
}

\maketitle
\section{Algorithm for finding maximum subpath in the 3D tensor}
We show the algorithm to find the maximum subpath in the 3D tensor based on the dynamic programming proposed by~\cite{tran2014video} as below.

\begin{algorithm}[h!]
\caption{Algorithm for finding maximum subpath in the 3D tensor~\cite{tran2014video}} 
\label{alg1}
\begin{framed}
\begin{algorithmic}
\REQUIRE
  \STATE{$M(u,t):$ the local discriminative scores;}
  \COMMENT{$u = (i,j):$ the 2D index of spatial coordinate}
  \COMMENT{$t:$ the frame in the video}  
\ENSURE
  \STATE{$S(u,t):$ the accumulated scores of the best path leads to $(u,t)$;}
  \STATE{$P(u,t):$ the best path record for tracing back;}
  \STATE{$S^* :$ the accumulated score of the best path;}
  \STATE{$l^* :$ the ending location of the best path;}
　\STATE{\\}
  \STATE{$S(u,1) = M(u,1), \forall u$;}
  \STATE{$P(u,t) = null, \forall (u,t);$}
  \STATE{$S^* = -\infty;$}
  \STATE{$l^* = null;$}\\
\FOR{$i \leftarrow 2$ \bf{to} $n$}
  \FORALL{$u \in [1..w] \times [1..h]$} 
    \STATE{$v_0 \leftarrow \argmax_{v \in N(u)} S(v,i-1);$}
    \IF{$S(v_0,i-1)>0$}
      \STATE{$S(u,i) \leftarrow S(v_0,i-1) + M(u,i);$}
      \STATE{$P(u,i) \leftarrow (v_0, i-1);$}
    \ELSE
      \STATE{$S(u,i) \leftarrow M(u,i);$}
    \ENDIF
    \IF{$S(u,i) > S^*$}
      \STATE{$S^* \leftarrow S(u,i);$}
      \STATE{$l^* \leftarrow (u,i);$}
    \ENDIF
  \ENDFOR
\ENDFOR
\end{algorithmic}
\end{framed}
\end{algorithm}

\newpage


\ 
\section{The influence of the complexity of classification model on the negative class accuracy}
\begin{figure*}[h!]
\centering
\begin{minipage}{.48\textwidth}
  \centering
  \includegraphics[width=.9\linewidth]{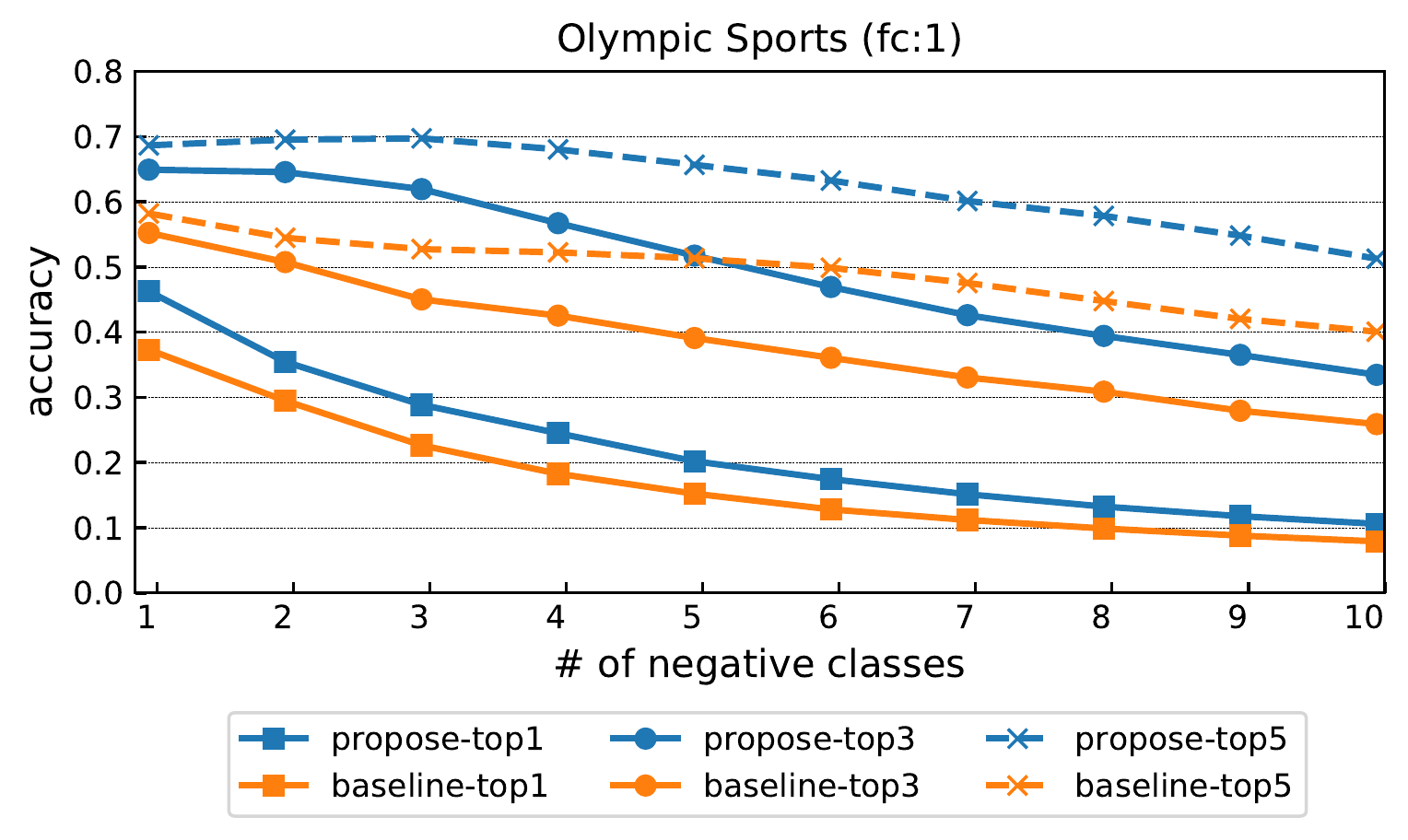}
\end{minipage}
\begin{minipage}{.48\textwidth}
  \centering
  \includegraphics[width=.9\linewidth]{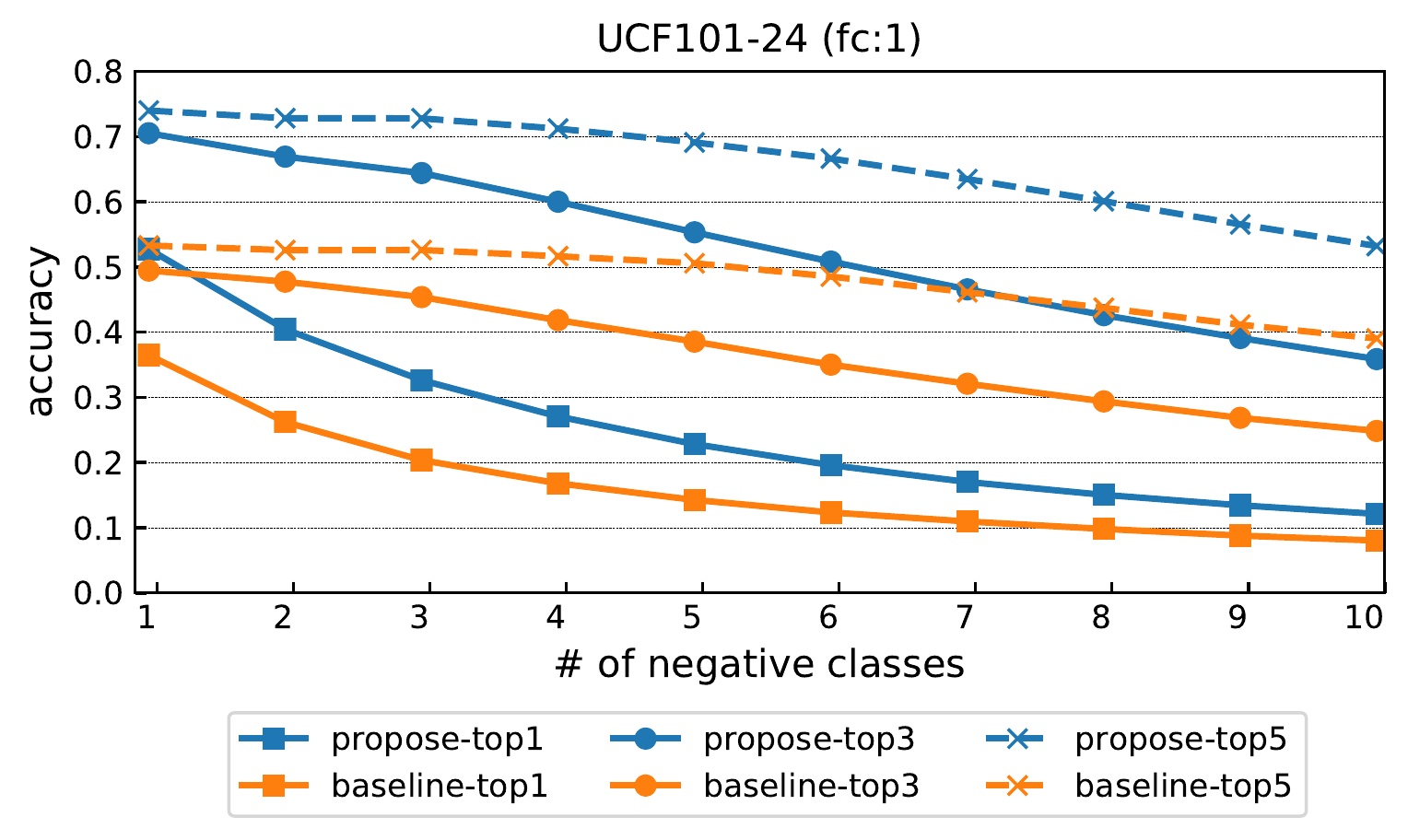}
\end{minipage}
\\
\begin{minipage}{.48\textwidth}
  \centering
  \includegraphics[width=.9\linewidth]{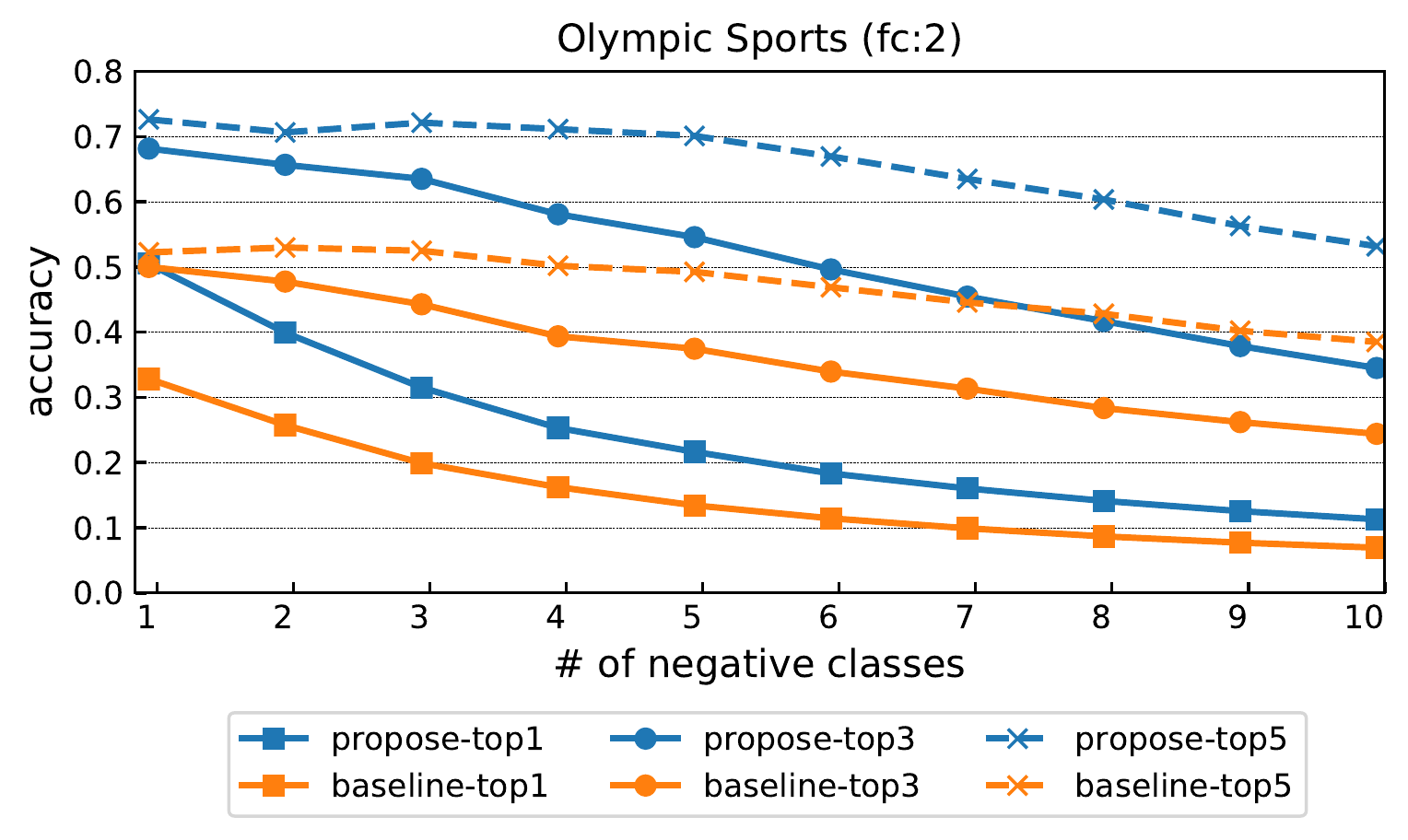}
\end{minipage}
\begin{minipage}{.48\textwidth}
  \centering
  \includegraphics[width=.9\linewidth]{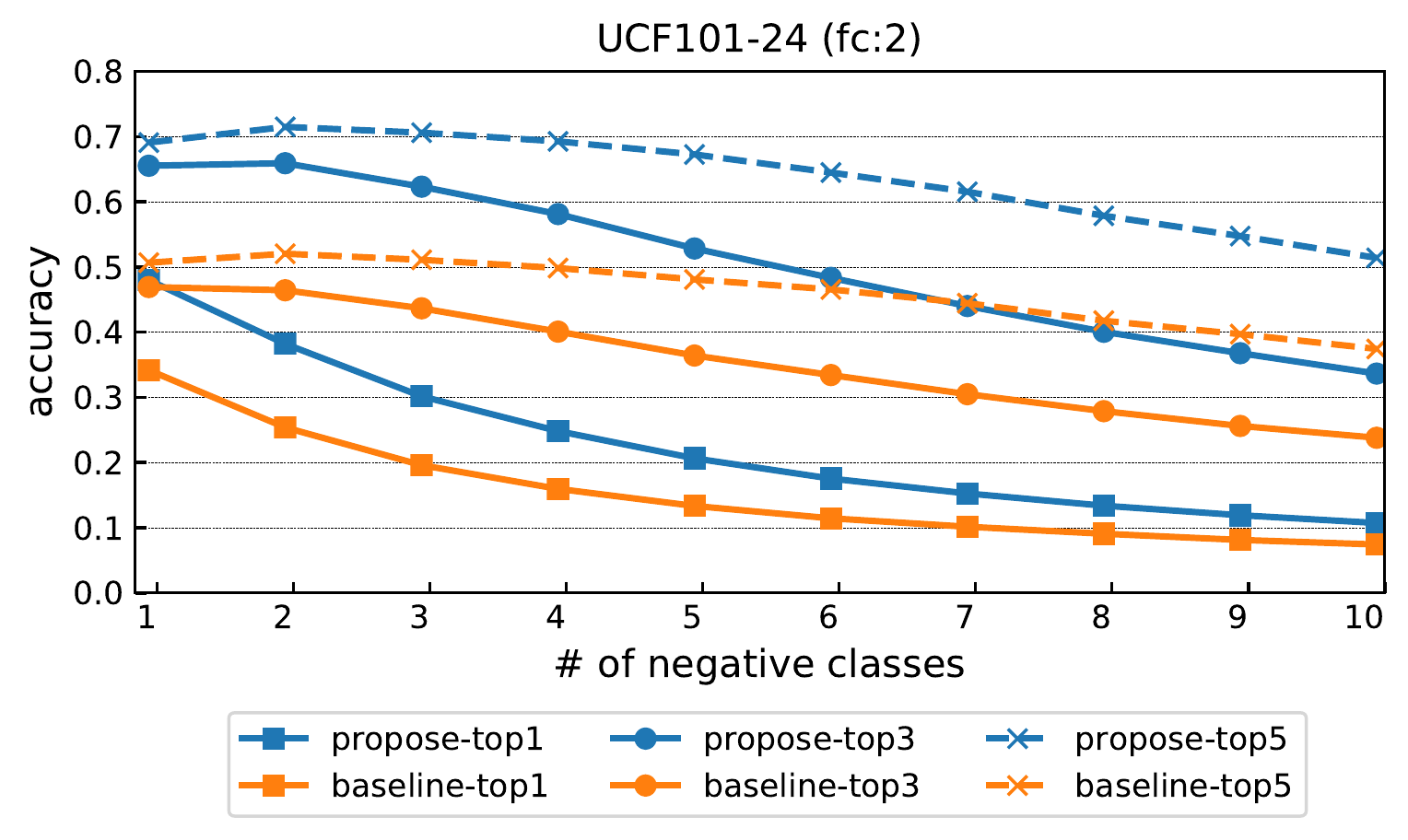}
\end{minipage}
\\
\begin{minipage}{.48\textwidth}
  \centering
  \includegraphics[width=.9\linewidth]{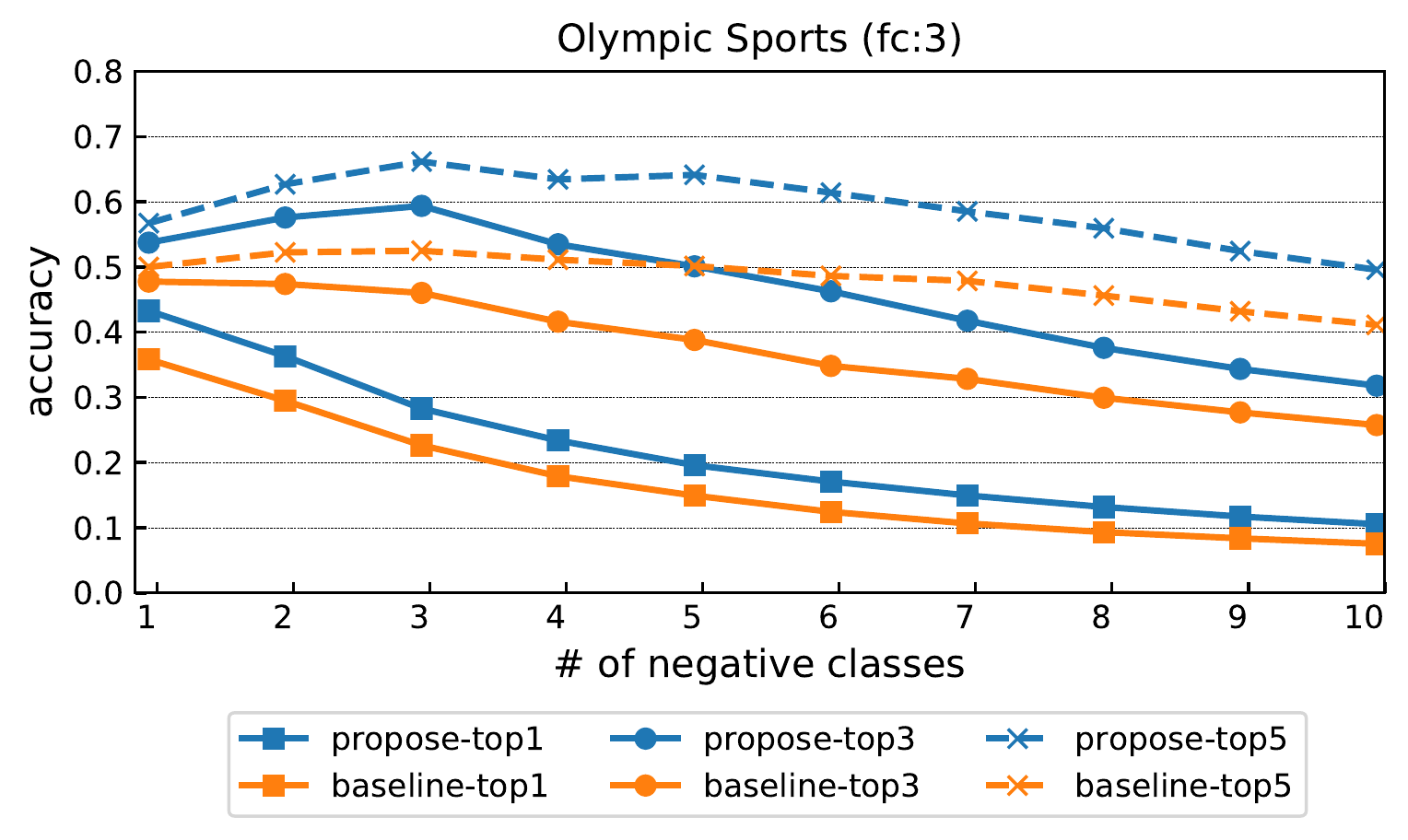}
\end{minipage}
\begin{minipage}{.48\textwidth}
  \centering
  \includegraphics[width=.9\linewidth]{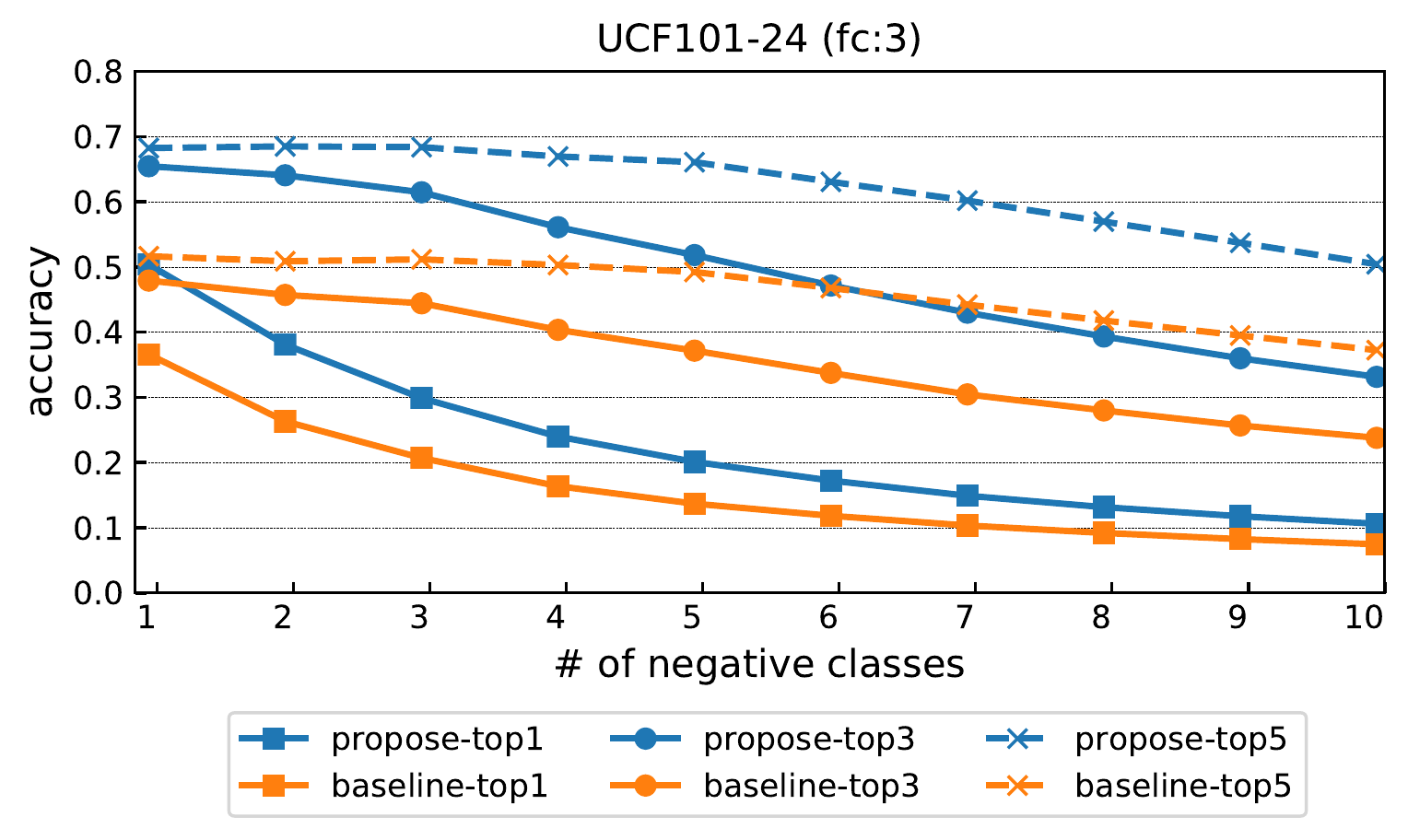}
\end{minipage}
\\
\vspace{1cm}
\caption{The negative class accuracy on Olympic Sports dataset (left) and UCF101-24 dataset (right). Each row corresponds to the number of fully-connected layer of the classification module. y-axis indicates the mean accuracy and x-axis means the number of the negative classes used for averaging whose prediction value is maximum.}
\label{fig:negacc}
\end{figure*}

\newpage
\section{List of attributes}
\subsection{Olympic Sports dataset}
'Run', 'Slow  run', 'Fast  run', 'Indoor', 'outdoor', 'Ball', 'small  ball', 'big  ball', 'Jump', 'Small  local  Jump', 'Local  jump  up', 'Jump  Forward', 'Track', 'Bend', 'StandUp', 'Lift  something', 'Raise  Arms', 'One  Arm  Open', 'Turn  Around', 'Throw  Up', 'Throw  away', 'water', 'Down  Motion  in  Air', 'Up  Motion  in  Air', 'Up  Down  Motion  Local', 'Somersault  in  Air', 'With  Pole', 'Two  hand  holding  pole', 'One  hand  holding  pole', 'Spring  Platform', 'Motion  in  the  air', 'one  arm  swing', 'Crouch', 'Two  Arms  Open', 'Two  Arms  Swing  overhead', 'Turn  around  with  two  arms  open', 'Run  in  Air', 'Big  Step', 'Open  Arm  Lift', 'With  Pat'
\subsection{UCF101-24 dataset}
'Body Motion is Flipping', 'Body Motion is Walking', 'Body Motion is Running', 'Body Motion is Riding', 'Body Motion is Up down', 'Body Motion is Pulling', 'Body  Motion is Lifting', 'Body Motion is Pushing', 'Body  Motion is Diving', 'Body Motion is Jumping  Up', 'Body  Motion is Jumping Forward', 'Body  Motion is Jumping  Over  Obstacle', 'Body  Motion is Spinning', 'Body  Motion is Climbing Up', 'Body Motion is Horizontal', 'Body  Motion is Vertical  Up', 'Body  Motion is Vertical  Down', 'Body  Motion is Bending', 'Object is Ball  Like', 'Object is Big  Ball  Like', 'Object is Stick  Like', 'Object is Rope  Like', 'Object is Sharp', 'Object is Circular', 'Object is Cylinderical', 'Object is Musical  Instrument', 'Object is Portable  Musical  Instrument', 'Object is Animal', 'Object is Boat  Like', 'Posture is Sitting', 'Posture is Sitting  In  Front  Of  Table  Like  Object', 'Posture is Standing', 'Posture is Lying', 'Posture is Handstand', 'Body  Parts  Used is Head', 'Body  Parts  Used is Hands', 'Body  Parts  Used is Arms', 'Body  Parts  Used is Legs', 'Body  Parts  Used is Foot'

\clearpage
\newpage 
\section{Output Examples}
\input{./img_supp_counter/readimg}

\newpage
\section{Dataset collection}

\begin{figure*}[h!]
\begin{center}
\begin{tabular}{c}
\includegraphics[clip, width=0.8\linewidth, height=0.85\textheight]{./img/beaverdam.pdf}
\end{tabular}
\end{center}
\caption{Screen shot of the instruction for collecting bounding box annotation on AWS.}
\label{fig:exp}
\end{figure*}

\newpage
\section{Dataset Examples}

\input{./img_supp_counter/readdataset}

\clearpage

{\small
\bibliographystyle{ieee}
\bibliography{egbib}
}